\documentclass[twoside,11pt]{article}

\usepackage{blindtext}

\usepackage[preprint]{jmlr2e}

\usepackage{color}
\usepackage{multirow}

\usepackage{lastpage}

\ShortHeadings{TinyLLaVA Factory}{Lei Huang and Ji Wu}
\firstpageno{1}

\begin{document}

\title{TinyLLaVA Factory: A Modularized Codebase for Small-scale Large Multimodal Models}

\author{\name Junlong Jia$^{*1}$, \ \name Ying \ Hu$^{*2}$, \ \name Xi \ Weng$^{*1}$, \ \name Yiming \ Shi$^{*2}$
	\AND
	\name Miao Li$^{2}$, \ \name Xingjian \ Zhang$^{1}$, \ \name Baichuan \ Zhou$^{1}$,  \ \name Ziyu \ Liu$^{2}$
	\AND
	\name Jie \ Luo$^{1}$, \ \name Lei \ Huang$^{\dag 1}$, \ \name Ji \ Wu$^{\dag 2}$ \\
	\AND
	\addr 1 \{jiajunlong, winci,  zhangxingjian, baichuanzhou, luojie, huangleiai\}@buaa.edu.cn \\
	\addr SKLCCSE, Institute of Artificial Intelligence\\
	Beihang University\\
	Beijing, China
	\AND
	\addr 2 \{yinghu\_yh, miao-li, wuji\_ee\}@mail.tsinghua.edu.cn \  \{sym23, ziyu-liu22\}@mails.tsinghua.edu.cn \\
	\addr Department of Electronic Engineering, College of AI\\
	Tsinghua University\\
	Beijing, China}

\let\thefootnote\relax\footnotetext{* means that Junlong Jia, Ying Hu, Xi Weng, and Yiming Shi contribute equally.}

\let\thefootnote\relax\footnotetext{$\dag$  means that Lei Huang and Ji Wu are corresponding authors.}

\maketitle

\begin{abstract}
We present TinyLLaVA Factory, an open-source modular codebase for small-scale large multimodal models (LMMs) with a focus on simplicity of code implementations, extensibility of new features, and reproducibility of training results. Following the design philosophy of the factory pattern in software engineering, TinyLLaVA Factory modularizes the entire system into interchangeable components, with each component integrating a suite of cutting-edge models and methods, meanwhile leaving room for extensions to more features. In addition to allowing users to customize their own LMMs, TinyLLaVA Factory provides popular training recipes to let users pretrain and finetune their models with less coding effort. Empirical experiments validate the effectiveness of our codebase. The goal of TinyLLaVA Factory is to assist researchers and practitioners in exploring the wide landscape of designing and training small-scale LMMs with affordable computational resources. \\
Code: \url{https://github.com/TinyLLaVA/TinyLLaVA_Factory}  \\
Documentation: \url{https://tinyllava-factory.readthedocs.io/en/latest/} \\
\end{abstract}


\vspace{-0.5em}
\section{Introduction}
\vspace{-0.5em}
Large Language Models (LLMs) have unified various language understanding and generation tasks~\citep{2020_NIPS_GPT-3}, by using auto-regressive prediction during training and instruction prompt during evaluation~\citep{2022_NIPS_InstructGPT}. This paradigm of task unification has spread to the computer vision community, giving rise to the Large Multimodal Models (LMMs) \citep{2021_NIPS_Frozen,2023_techreport_OpenFlamingo,2023_NIPS_LLaVA,2023_arxiv_minigpt-4} that treat visual inputs as conditional information and leverage powerful abilities of LLMs. While the unification of LMMs in modeling various visual and linguistic tasks shows great potential in building Artificial General Intelligence (AGI), the training of LMMs gets quite complicated for practitioners - it requires miscellaneous data preprocessing and careful collaboration between model architectures and training recipes. Furthermore, the scaling up of model sizes of LMMs requires expensive computational resources and leads to unaffordable training and evaluation budget, which restricts research access to only well-funded industries and organizations~\citep{2024_arxiv_TinyLLava}.

To address the above issues, this paper presents TinyLLaVA Factory, a modularized codebase with standard training$\&$evaluating pipelines, flexible data preprocessing$\&$model configurations, and easily extensible architectures. TinyLLaVA Factory adheres to the design philosophy of the factory pattern in software engineering, which modularizes the entire system into interchangeable components, with each component integrating a suite of cutting-edge models and methods. Keeping in line with factory pattern can decompose complicated model architecture and training process, thus enabling users to easily build their own LMMs with minimal coding effort and to prevent coding mistakes. It also features standard data-preprocessing pipelines and  popular training recipes, providing friendly interfaces for user customization. 

For small-scale LMMs that can be trained with limited computational resources, Tiny-LLaVA Factory integrates popular small-scale LLMs ranging from 450M to 2.7B, such as OpenELM-450M~\citep{mehta2024openelm}, TinyLlama-1.1B~\citep{2024_arxiv_TinyLlama}, StableLM-2-1.6B~\citep{2023_techreport_StableLM}, Gemma-2B~\citep{team2024gemma}, Phi-2-2.7B~\citep{2023_techreport_Phi2}. Despite the fact that TinyLLaVA Factory targets at small-scale LMMs, the codebase is applicable to large-scale LMMs by simply scaling up LLMs. Finally, we provide empirical experiments using this codebase among different small-scale LMMs. 
\section{Architecture and Key Components}
\begin{figure}[t]
        \vspace{-0.2in}
	\centering
	{\includegraphics[width=15.5cm]{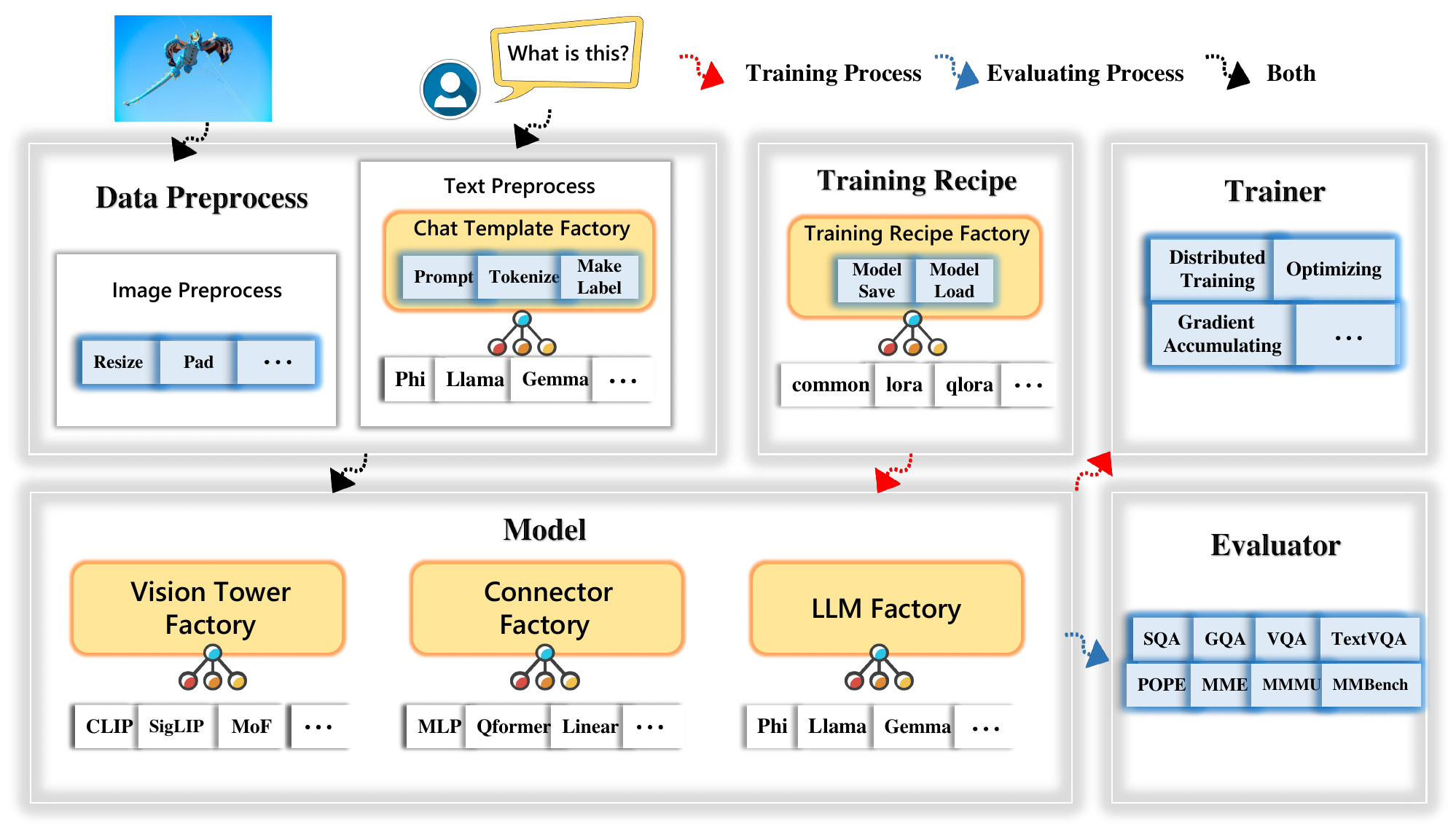} }
        \vspace{-0.25in}
	\caption{The overall architecture of TinyLLaVA Factory.}
	\label{fig:architecture}
	\vspace{-0.03in}
\end{figure}

TinyLLaVA Factory follows software principles of modularity, simplicity, and extensibility to build an ecosystem for training and evaluating small-scale LMMs, implemented in PyTorch~\citep{paszke2019pytorch} and Hugging Face and equipped with DeepSpeed~\citep{rasley2020deepspeed}. The overall architecture of TinyLLaVA Factory is illustrated in Figure~\ref{fig:architecture}. It leverages the standard deep learning pipeline that starts from preparing data, preparing model, to training and evaluating. TinyLLaVA Factory is broken down into five modules that can be interchangeable and manageable: data, model, training recipe, trainer, and evaluator. We endeavor to minimize the dependencies between these modules, allowing each module to be scaled to new alternatives in a plug-and-play fashion. Specifically, data, model, and training recipe are managed by factory pattern that is responsible for registering new models or training recipes. A factory goes hand-in-hand with a base class that offers basic and universal properties and functions. Users’ customized modules can inherit from the base class with minimal code changes. Note that though TinyLLaVA Factory aims to train small-scale LMMs, the codebase is applicable to large-scale LMMs by simply scaling up LLMs.
\vspace{-0.1em}
\subsection{Key Components}
\noindent \textbf{Data.} The datasets that are used for pretraining and finetuning should follow LLaVA~\citep{2023_NIPS_LLaVA} data format. Data processing includes image and text preprocessing. For image preprocessing, typical operations like resizing, cropping, and normalizing are performed. For text preprocessing, chat template is important because it is usually related to the instruction following ability of models. Therefore, we design a base data class of chat template providing functions of prompting text using system message, tokenizing text, and making labels for ground truth answers. We provide subclasses of chat templates for popular LLMs like Llama, Phi, and Gemma, which are inherited from the base class.

\noindent \textbf{Model.} The model part of TinyLLaVA Factory is further modularized as three components: small-scale LLM, vision tower, and intermediate connector. Each component is associated with a factory for registering new models that can be created based on the base class. For LLM and vision tower, we have provided a suite of alternative cutting-edge models and allow users to easily replace these models by simply specifying configuration files. For example, vision tower supports Openai CLIP ViT, Google SigLIP ViT, Meta Dinov2, or MoF combining both CLIP and Dinov2. LLM supports Phi, Gemma, OpenELM, among others. (Table 2 in Appendix lists models and methods that have been implemented in TinyLLaVA Factory.)

\noindent \textbf{Training Recipe.} Existing work ~\citep{2023_arxiv_mPlug-Owl, 2023_NIPS_LLaVA, 2023_arxiv_MiniGPT-v2} favors in adopting multi-stage training procedures where the LLM, vision tower, and connector component can be either frozen, fully tuned, or LoRA tuned in different stages. TinyLLaVA Factory designs a class named training recipe to control the tuning type of each component. Users can customize their own training recipes for more complex needs by inheriting from the base training recipe. Furthermore, the design of training recipes facilitates the selection of appropriate methods for saving and loading models according to tuning type and DeepSpeed mode.

\noindent \textbf{Trainer.} After specifying LMM components and training recipe, TinyLLaVA Factory leverages a Hugging Face trainer for feature-complete training. TinyLLaVA training is powered by features that are built in Hugging Face, including gradient accumulation, DeepSpeed ZeRO, report logging to TensorBoard or Wandb.

\noindent \textbf{Evaluation, Testing, and Documentation.}  TinyLLaVA Factory currently provides evaluations on 8 benchmarks, including SQA, GQA, POPE, MMMU, and more. For testing, we conduct integration tests for training and evaluating across multiple modules to verify that they function correctly together with a coverage of around 92\% of code lines, which guarantees the quality and executability of the code. For documentation, we provide installation instructions and descriptions of methods and functions and their parameters and returned values.


\vspace{-0.5em}
\subsection{Comparison to Related Codebase}
TinyLLaVA Factory is related to the released codebase of LLaVA~\citep{2023_NIPS_LLaVA}, but has several merits: 1) Our codebase is modularized with standard training$\&$evaluating pipelines and flexible data preprocessing$\&$model configurations, while the codebase of LLaVA neglects these characteristics from the perspective of software designing; 2) LLaVA treats vision tower and connector as the property of a LLM, while our TinyLLaVA Factory views LLM, vision tower and connector as the components of the LMMs, which is more extensible to integrate more capabilities, such as adding the vision generation component. We also note that the codebase released by prismatic-vlms~\citep{2024_arxiv_PrismaticVLM} uses this design for LMMs and enjoys the extensibility. Different from Prismatic-VLMs, our TinyLLaVA Factory adopts the design philosophy of the factory pattern in software engineering, and modularizes the entire system into interchangeable components, with each component already integrating a suite of cutting-edge models,  standard data-processing pipelines, and  popular training recipes.

\setlength{\tabcolsep}{3pt}
\begin{table*}[t]
\caption{The reproduction results of TinyLLaVA variants.  "VT", "LLM", and "Recipe" respectively represent Vision Tower, small-scale Large Language Model, and Training recipe.}
    \begin{footnotesize}
    \centering
\begin{tabular}{c|c|c|cccc|cccc}
\hline
VT & LLM & Recipe &
 \multicolumn{4}{c}{Image Question Answering} & \multicolumn{4}{c}{Benchmark Toolkit} \\
   & & &   VQA$^{v2}$ & GQA &  SQA$^I$ & VQA$^T$ & MM-Vet & POPE & MME & MMMU-val\\

\hline
CLIP    & OpenELM-450M   & base   & 69.5 & 52.1  & 50.6  & 40.4  & 20.0     & 83.6  & 1052.9 & 23.9\\
SigLIP    & OpenELM-450M   & base   & 71.7 & 53.9  & 54.1  & 44.0  & 20.0     & 85.4  & 1118.8 & 24.0\\
CLIP    & TinyLlama-1.1B   & base   & 73.7 & 58.0  & 59.9  & 46.3  & 23.2     & 85.5  & 1284.6 & 27.9  \\
SigLIP    & TinyLlama-1.1B   & base   & 75.5 & 58.6  & 64.0  & 49.6  & 23.5     & 86.3  & 1256.5 & 28.3 \\
CLIP    & StableLM-2-1.6B   & base & 75.9 & 59.5  & 64.6  & 50.5  & 27.3     & 86.1   & 1368.1 & 31.8 \\
SigLIP   & StableLM-2-1.6B   & base  & 78.2 & 60.7  & 66.7  & 56.0  & 29.4     & 86.3  & 1319.3 & 32.6 \\
SigLIP   & Gemma-2B  & base  & 78.4 & 61.6  & 64.4 & 53.6  & 26.9     & 86.4  & 1339.0 & 31.7\\
CLIP   & Phi-2-2.7B   & base & 76.8 & 59.4  & 71.2  & 53.4  & 31.7     & 86.8 & 1448.6 
 & 36.3\\
SigLIP   & Phi-2-2.7B   & base & 79.2 & 61.6  & 71.9  & 57.4  & 35.0     & 87.2   & 1462.4 & 38.2\\
SigLIP   & Phi-2-2.7B   & share  & 80.1 & 62.1  & 73.0  & 60.3  & 37.5  & 87.2 & 1466.4 & 38.4 \\

 \hline
\end{tabular}
\end{footnotesize}
\label{tab:repro}
\end{table*}

\section{Experiments}
We conduct experiments (see detailed settings in Appendix~\ref{appendix:a}) to reproduce several variants of TinyLLaVA and assess their performance on standard benchmarks within our codebase, as outlined in Table~\ref{tab:repro}, achieving slightly superior overall performance compared to the performance reported in the original paper~\citep{2024_arxiv_TinyLLava} that were trained with the LLaVA codebase. These results highlight the reliability of our TinyLLaVA Factory and offer valuable insights into the performance of TinyLLaVA.

\section{Conclusion and Future Work}
To facilitate the open research on small-scale LMMs, we introduce TinyLLaVA Factory, an open-source codebase implemented in Pytorch and Hugging Face for training small-scale LMMs, adhering to the design philosophy of modularity, simplicity, and extensibility, meanwhile guaranteeing the reproducibility of training results. In the future, we will integrate more efficient fine-tuning techniques and will continuously keep up to date with state-of-the-art models. We also encourage contributions from the open-source community.

\paragraph{Acknowledgment.}
This work was partially supported  by the National Key Research and Development Plan of China under Grant 2022ZD0116310, National Natural Science Foundation of China (Grant No. 62106012), the Fundamental Research Funds for the Central Universities.

\vskip 0.2in
\bibliography{FoudationModel}

\newpage

\begin{appendix}
\setlength{\tabcolsep}{2pt}

\section{Experimental Settings}
\label{appendix:a}
In Table~\ref{tab:repro}, we conduct extensive experiments to evaluate various models and key components within our codebase. The specific experimental settings are described as follows.

\paragraph{Models.} We reproduce all the model types used in the original TinyLLaVA paper, including small-scale LLMs (TinyLlama-1.1B, StableLM-2-1.6B, and Phi-2-2.7B), Vision Towers (OpenAI CLIP ViT and Google SigLIP ViT), and the connector (a two-layer MLP with GELU activation). Additionally, we implement the language models OpenELM-450M and Gemma-2B in the codebase.

\paragraph{Training Recipes.} We follow the training recipes (\textbf{base} and \textbf{share}) used in the original TinyLLaVA paper. Specifically, for the \textbf{base} recipe, during pre-training, only the connector is updated while the rest of the model remains frozen. The model is trained for one epoch with a learning rate of 1e-3 and a batch size of 256. In the supervised fine-tuning stage, the vision tower is kept frozen while both the connector and the small-scale LLM are updated. The model is tuned for one epoch with a learning rate of 2e-5 and a batch size of 128. 

During pre-training of the \textbf{share} recipe, the connector is initialized from the base's pretrained counterpart. Additionally, the vision tower is kept frozen while the rest of the model is updated for one epoch with a learning rate of 2e-5 and a batch size of 256. The setup for supervised fine-tuning is the same as the \textbf{base} recipe. 

\paragraph{Training Data and Data Preprocessing.}
By default, the LLaVA-1.5 dataset is utilized for training when using the \textbf{base} recipe, whereas the ShareGPT4V dataset is employed for training with the \textbf{share} recipe. We follow the data preprocessing settings outlined in the original TinyLLaVA paper, with the exception that we configure the \textit{image\_aspect\_ratio} parameter to square in both pretraining and finetuning stage when using SigLIP. This adjustment leads to slightly superior performance compared to the results reported in the original paper. 

\paragraph{Evaluating Benchmarks.}

We evaluate our reproduced models on four image question-answering benchmarks: VQA-v2~\citep{2017_CVPR_vqav2}, GQA~\citep{2019_CVPR_GQA}, ScienceQA-IMG~\citep{2022_NIPS_SQA}, and TextVQA~\citep{2019_CVPR_TextVQA}, and four comprehensive benchmark:  MM-Vet~\citep{2023_arxiv_MMVet}, POPE~\citep{2023_EMNLP_POPE}, MME~\citep{2023_arxiv_MME}, and MMMU~\citep{2023_arixv_MMMU}.
\end{appendix}

\begin{table*}[t]
   \caption{ Small-scale LLMs, vision towers, connectors, and training recipes that have been implemented in TinyLLaVA Factory.}
 	\vspace{0in}
 	\centering
 	\setlength{\tabcolsep}{4pt}
 	\begin{tabular}{p{1.75cm}|p{6.6cm}||p{1.3cm}|p{5cm}}
 		\hline
 		Module & Name & Module & Name  \\
 		\hline
		\multirow{6}{2.2cm}{Small-scale LLM} & OpenELM-450M~\citep{mehta2024openelm} & \multirow{6}{1.5cm}{Vision\\Tower} & \multirow{6}{5cm}{CLIP~\citep{2021_ICML_CLIP} \\ SigLIP~\citep{2023_ICCV_SigLIP} \\ DinoV2~\citep{oquab2023dinov2} \\ MoF~\citep{2024_arxiv_mof} }\\
             & TinyLlama-1.1B~\citep{2024_arxiv_TinyLlama}  & & \\
 		& StableLM-2-1.6B~\citep{2023_techreport_StableLM}  & & \\
             & Qwen-1.5-1.8B~\citep{2024_techreport_Qwen1.5} & & \\
             & Gemma-2B~\citep{team2024gemma}  &  & \\
 		& Phi-2-2.7B~\citep{2023_techreport_Phi2} & & \\  
 		\hline
 		\hline
		\multirow{5}{*}{Connector} & Identity & \multirow{5}{2.2cm}{Training Recipe} & \multirow{5}{5cm}{Frozen/Full/Partially Tune \\ LoRA/QLoRA~\citep{hu2021lora, dettmers2024qlora} }\\
		& Linear &  &  \\
             & MLP &  &  \\
             & Q-former~\citep{2023_arxiv_BLIP2} &    \\
             & Resampler~\citep{2023_techreport_OpenFlamingo} &  &  \\
 		\hline
 	\end{tabular}
 	\setlength{\tabcolsep}{5pt}
 	\label{tab:models}
 \end{table*}

\end{document}